\documentclass{article}

\usepackage{PRIMEarxiv}

\usepackage[utf8]{inputenc} 
\usepackage[T1]{fontenc}    

\usepackage{hyperref}
\usepackage{amsmath,amssymb,amsfonts}
\usepackage{tabularx}
\usepackage{subfigure}
\usepackage{graphicx}
\usepackage{cite}
\usepackage[linesnumbered,ruled,vlined]{algorithm2e}
\usepackage{bm}
\usepackage{booktabs}
\usepackage{caption}

\title{\LARGE \bf
Dynamic Hand Gesture-Featured Human Motor Adaptation in Tool Delivery using Voice Recognition\thanks{This work has been submitted to the IEEE for possible publication. Copyright may be transferred without notice, after which this version may no longer be accessible.}}

\author{Haolin Fei$^{1}$, Stefano Tedeschi$^{2}$, Yanpei Huang$^{3}$, Andrew Kennedy$^{1}$, and Ziwei Wang$^{1}$
\thanks{$^{1}$Haolin Fei, Andrew Kennedy, and Ziwei Wang are with the School of Engineering, Lancaster University, Lancaster, UK. $^{2}$Stefano Tedeschi is with the Centre for Digital Engineering and Manufacturing, School of Aerospace, Transport and Manufacturing, Cranfield University, Cranfield, UK. $^{3}$Yanpei Huang is with the Department of Bioengineering, Imperial College of Science, Technology and Medicine, London, UK. 
\{h.fei1, z.wang82\}@lancaster.ac.uk}
}

\begin{document}
\maketitle
\begin{abstract}
Human-robot collaboration has benefited users with higher efficiency towards interactive tasks. Nevertheless, most collaborative schemes rely on complicated human-machine interfaces, which might lack the requisite intuitiveness compared with natural limb control. We also expect to understand human intent with low training data requirements. In response to these challenges, this paper introduces an innovative human-robot collaborative framework that seamlessly integrates hand gesture and dynamic movement recognition, voice recognition, and a switchable control adaptation strategy. These modules provide a user-friendly approach that enables the robot to deliver the tools as per user need, especially when the user is working with both hands. Therefore, users can focus on their task execution without additional training in the use of human-machine interfaces, while the robot interprets their intuitive gestures. The proposed multimodal interaction framework is executed in the UR5e robot platform equipped with a RealSense D435i camera, and the effectiveness is assessed through a soldering circuit board task. The experiment results have demonstrated superior performance in hand gesture recognition, where the static hand gesture recognition module achieves an accuracy of 94.3\%, while the dynamic motion recognition module reaches 97.6\% accuracy. Compared with human solo manipulation, the proposed approach facilitates higher efficiency tool delivery, without significantly distracting from human intents.
\end{abstract}


\section{Introduction}
Human-robot collaboration (HRC) has offered a compelling synergy between humans and machines in various domains\cite{Jacopo2023Towards}, where the need for efficient and seamless interactions between humans and robots becomes increasingly imperative \cite{sheridan2016human}. Hand-gesture-based HRC presents a range of distinct advantages in human-robot collaboration. Firstly, it capitalizes on a mode of communication that comes instinctively to humans. Hand gestures constitute an integral part of everyday interactions, necessitating no specialized equipment or training, thereby reducing barriers to entry for users. Secondly, hand gestures facilitate non-verbal communication, which can be particularly advantageous in noisy or crowded environments where voice commands might prove less effective. They can also provide an additional layer of communication, allowing users to convey nuanced instructions and preferences beyond what can be expressed through words alone.

\begin{figure}[t]
\centering
\includegraphics[width=1\linewidth]{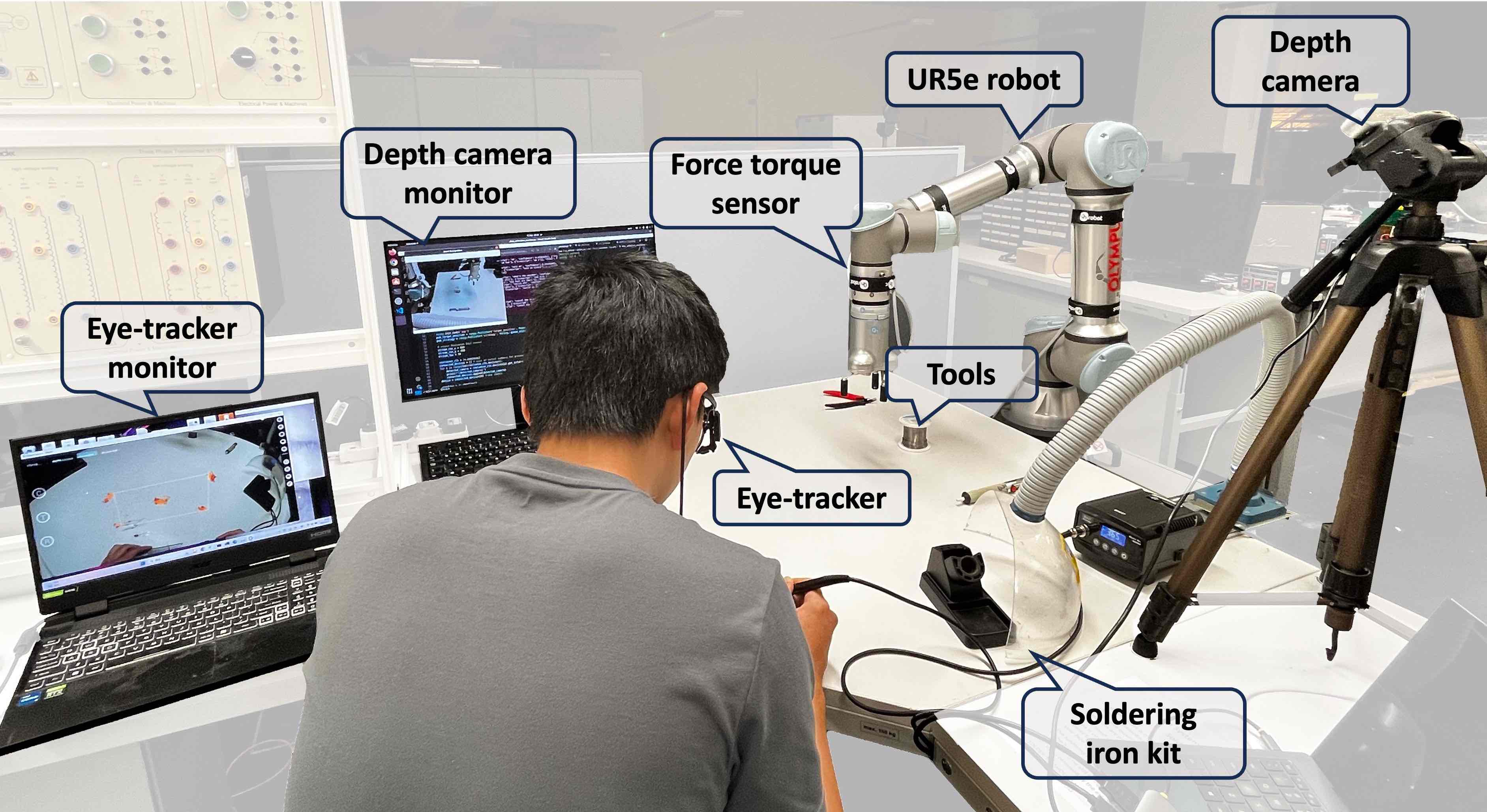}
\caption{A practical validation platform designed to assess multimodal interaction during the electrical circuit repair handover task.}
\label{fig_setup}
\end{figure}

However, existing hand gesture-based HRC systems predominantly revolve around task-based processes \cite{mazhar2019real,peral2022efficient,ju2015integrative}, often overlooking the human factors. While these systems excel at deciphering specific gestures to trigger predefined actions, they frequently fall short of comprehending the broader context and the nuanced needs of the human user. Moreover, many of these systems rely on overt and exaggerated hand gestures \cite{neto2019gesture,nuzzi2021hands,xu2015online}, which can be unnatural and fatiguing for users, especially over extended periods. Such gestures may also lack the subtlety required for conveying intricate instructions or preferences effectively. 

Motivated by the aforementioned challenges, this paper presents a novel HRC framework that demonstrates exceptional utility when both hands of users are engaged in tasks, rendering the capacity to issue commands without concern for the robot's method of execution of paramount importance. This interaction paradigm aligns with the concept of the \textit{supernumerary limb}, which allows users to extend their control through extra limbs or tools to manipulate the environment or interact with objects \cite{eden2022principles}. In this context, users can focus on their task execution, with the robot adeptly interpreting their intuitive gestures.
\begin{figure*}[t]
\centering
\includegraphics[width=1.0\linewidth]{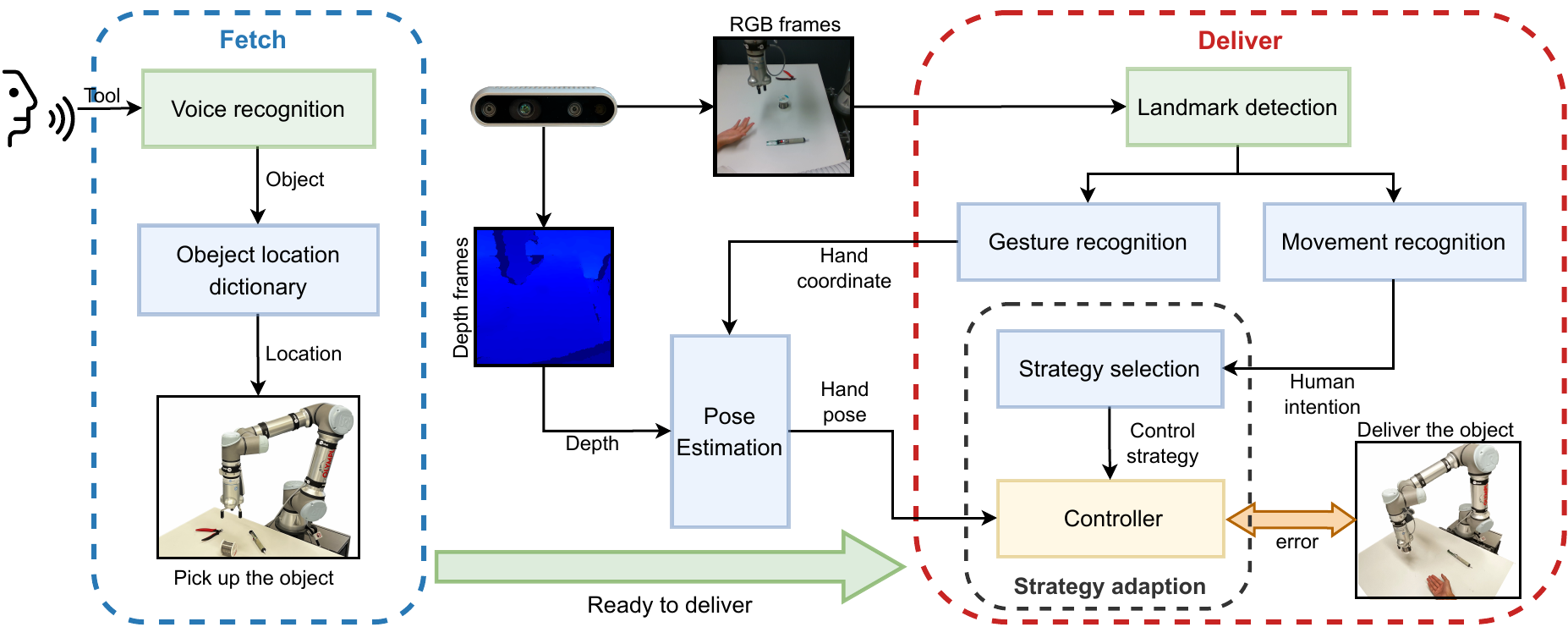}
\caption{Schematic representation of the comprehensive framework for HRC in dynamic tool delivery. The framework encompasses two fundamental stages: robot fetching and tool delivery. In the robot-fetching stage, voice command recognition enables users to specify desired tools verbally. The robot employs visual feedback to recognize and track the user's hand, estimating its 3D pose and discerning the user's intention. In the tool-delivering stage, real-time hand pose estimation through a depth camera ensures precise tool delivery.}
\label{framework}
\end{figure*}
To this end, our framework capitalizes on the rich visual cues embedded within human hand gestures. By analyzing both the form and dynamics of hand gestures, our system not only discerns the intended action but also gauges the degree of urgency or precision required by the user. This approach empowers the robot to adapt dynamically to the user's needs, delivering tools or objects with a level of finesse that was previously challenging to attain. Furthermore, we integrate voice command capabilities into the interaction system, allowing users to effortlessly and naturally instruct the robot. With this voice-command interface, users do not need to undergo extensive training or adapt to a rigid set of predefined gestures. In this way, users can interact with the robot without additional human-machine interfaces such as screens or controllers. The main contributions of this paper are presented as follows.

\begin{itemize}
    \item We introduce an HRC system that utilizes hand gesture-based information to interpret human intents, allowing the robot to dynamically adapt control strategies. This prioritizes understanding human intents and preferences, rendering a natural, user-friendly approach without additional training on using human-machine interfaces.
    \item We leverage hand gesture pose information in conjunction with voice commands, enabling robots to dynamically adjust to user requirements and confidently deliver tools and objects, regardless of variations in hand position.
    \item A multimodal verification platform has been developed involving soldered circuit board tasks that allow real-time monitoring of the user's workload throughout the interaction.
\end{itemize}

\section{Related Works}
One of the challenges faced by HRC is designing robots that can collaborate with humans in a natural and intuitive way. Humans have evolved to interact with other humans, and designing robots that can interact with humans in a way that feels natural and intuitive is a difficult task. Recent developments in HRC have pursued two primary approaches to tackle this challenge: predictive human motion modeling \cite{Wang2023Learning, martinez2017human, awais2012proactive} and intuitive interaction modalities \cite{li2022review}. 

Predictive human motion modeling leverages machine learning and predictive modeling techniques to enable robots to proactively adapt to human behavior. This approach aims to create interactions that are more anticipatory and responsive, enhancing the sense of naturalness in HRC. Traditionally, understanding and fulfilling human intentions or task requirements from the robot's perspective have relied on additional sensors and interfaces. However, these conventional methods come with their own set of limitations. One notable issue is the reliance on additional devices and signal processing, including technologies such as electromyography (EMG) \cite{ison2015proportional, su2021deep, nowak2016let, yang2016development}, electroencephalography (EEG) \cite{buerkle2021eeg}, or physiological signal-based approaches \cite{kim2020heterogeneous, kahanowich2021robust, Cheng2023Foot}. While these techniques have shown promise in certain contexts, they introduce elevated costs and increased complexity into the interaction setup. Users are often required to wear or employ these devices, which can be cumbersome and detract from the natural flow of interaction. Furthermore, large datasets are often required for training in human-robot interaction \cite{kim2017intrinsic}, which can be time-consuming and resource-intensive. This process, although crucial for the machine to understand human intentions, may not capture the full range of behaviors. Consequently, these systems may struggle to adapt beyond their training data, limiting their versatility in real-world settings. Thus, there's a need for more intuitive, adaptable, and less cumbersome approaches to HRC that align better with natural human communication and require minimal additional equipment or extensive data collection.

\begin{figure*}[t]
\centering
\includegraphics[width=1.0\linewidth]{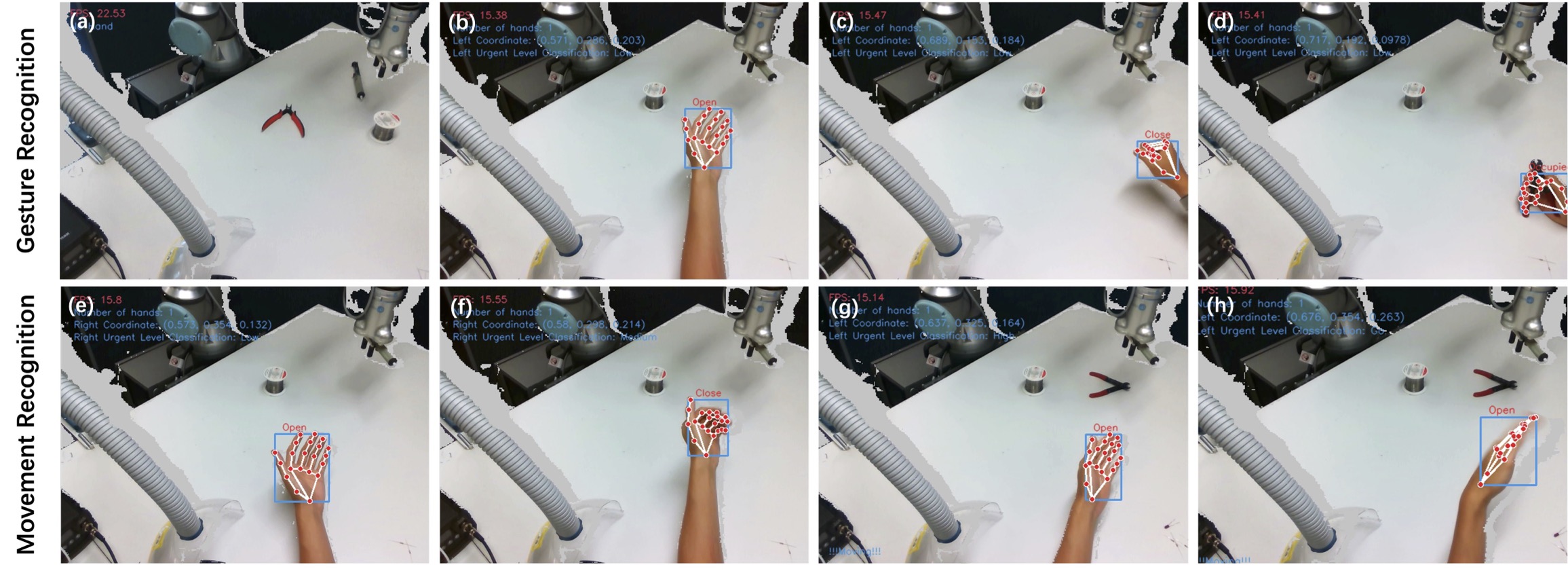}
\caption{Sample results from gesture and hand movement recognition frames, illustrating various scenarios. To enhance clarity, depth and RGB images have been combined, with pixels corresponding to point cloud data beyond the defined range omitted. (a) No hand. (b) Open hand gesture. (c) Closed hand gesture. (d) Occupied hand gesture. (e) Low urgency hand movement. (f) Medium-urgency hand movement. (g) High-urgency hand movement. (h) `Go away' hand movement.}
\label{gesture}
\end{figure*}

In addition, the other branch emphasizes intuitive interaction modalities, exemplified by technologies such as Virtual Reality (VR) \cite{malik2020virtual}, Augmented Reality (AR) \cite{michalos2016augmented}, Mixed Reality (MR) \cite{frank2016realizing}, voice recognition \cite{brock2021personalization}, hand gesture recognition \cite{mazhar2019real,peral2022efficient}. These modalities enable robots to communicate with humans in ways that closely resemble natural human interaction patterns, fostering greater user adaptability and improving the overall HRC experience.

In contrast to them, our approach integrates both concepts to forge effective HRC. We not only harness predictive modeling to enable robots to anticipate and adapt to human motion but also employ intuitive interaction modalities, including hand gesture recognition and voice commands. This comprehensive approach places the human user's needs at the forefront, culminating in a more natural and adaptable collaboration between humans and robots.




\section{Method}

\subsection{Framework Overview}
Our human-robot collaborated tool delivery framework is structured into two stages: the robot fetching and the tool delivering, each playing a crucial role in the overall process. Fig.\, \ref{framework} provides a visual representation of this framework.


In the initial phase, our primary aim is to establish a seamless and intuitive mode of communication between the human user and the robot. To this end, we deploy voice command recognition, enabling users to verbally specify their desired tool. We harness Google's Speech Recognition \cite{zhang2017speechrecognition} technology to convert spoken commands into actionable instructions for the robot. Following command processing, the robot initiates the tool retrieval process and proceeds to fetch the requested item. Subsequently, it awaits further directives from the human user.

In the subsequent tool-delivering stage, our emphasis shifts to the robot's capacity to perceive and respond to the user's intentions. To begin, we assess whether the human hand is within the robot's field of view. Upon detection, we proceed to extract key human hand landmarks and subsequently pass this information to a hand gesture recognition network. This network determines if the human hand's gesture indicates readiness to receive an object.

\subsection{Hand Pose Estimation}


We implement real-time hand pose estimation through a depth camera. This technology precisely identifies the exact position of the human hand's palm center within the robot's base frame. Transforming pixel coordinates into the robot's base frame involves a series of crucial steps. Firstly, we deproject pixel coordinates (\(u\) and \(v\)) to 3D Cartesian coordinates (\(X\), \(Y\), and \(Z\)) in the camera frame using intrinsic parameters:

\begin{equation}
    P_c = \begin{bmatrix}X \\ Y \\ Z\end{bmatrix} = D \cdot \begin{bmatrix}\frac{u - u_c}{f_x} \\ \frac{v - v_c}{f_y} \\ 1\end{bmatrix},
\end{equation} 
where \(P_c\) represents the 3D point in the camera frame, \(f_x\) and \(f_y\) represent the camera's focal lengths, \(u_c\) and \(v_c\) represent the camera's principal points, and \(D\) is the depth value obtained from the camera. Following this pixel-to-3D point conversion, precise calibration between the camera's reference frame and the robot's end-effector frame is established. This calibration, known as ``eye-to-hand calibration," accounts for any misalignments or offsets between the two frames. The transformation matrix \(T_{\text{eye-to-hand}}\) is computed to convert the 3D points from the camera frame to the robot's end-effector frame. Lastly, we translate the 3D point from the robot's end-effector frame to its base frame, which is represented by \(T_{\text{end-effector-to-base}}\). The overall transformation from pixel coordinates to the 3D point in the robot's base frame \(P_{\text{base}}\) is defined as:
\begin{equation}
    P_{\text{base}} = T_{\text{end-effector-to-base}} \cdot T_{\text{eye-to-hand}} \cdot P_c.
\end{equation}


\subsection{Learn to Collaborate from Hand Gesture }
We initiate our collaborative learning process by leveraging the highly efficient Mediapipe pose detector, a tool developed by Google \cite{lugaresi2019mediapipe}. Renowned for its proficiency, this framework employs cutting-edge machine learning techniques to precisely identify and locate specific landmarks on the human hand. By integrating the Mediapipe pose detector into our framework, we not only enhance interaction efficiency but also simplify our approach significantly. This streamlined approach not only improves training efficiency but also reduces computational demands. Additionally, it lessens the need for extensive training data, a common requirement when working directly with raw images. 


\begin{figure*}[t]
\centering
\includegraphics[width=1.0\linewidth]{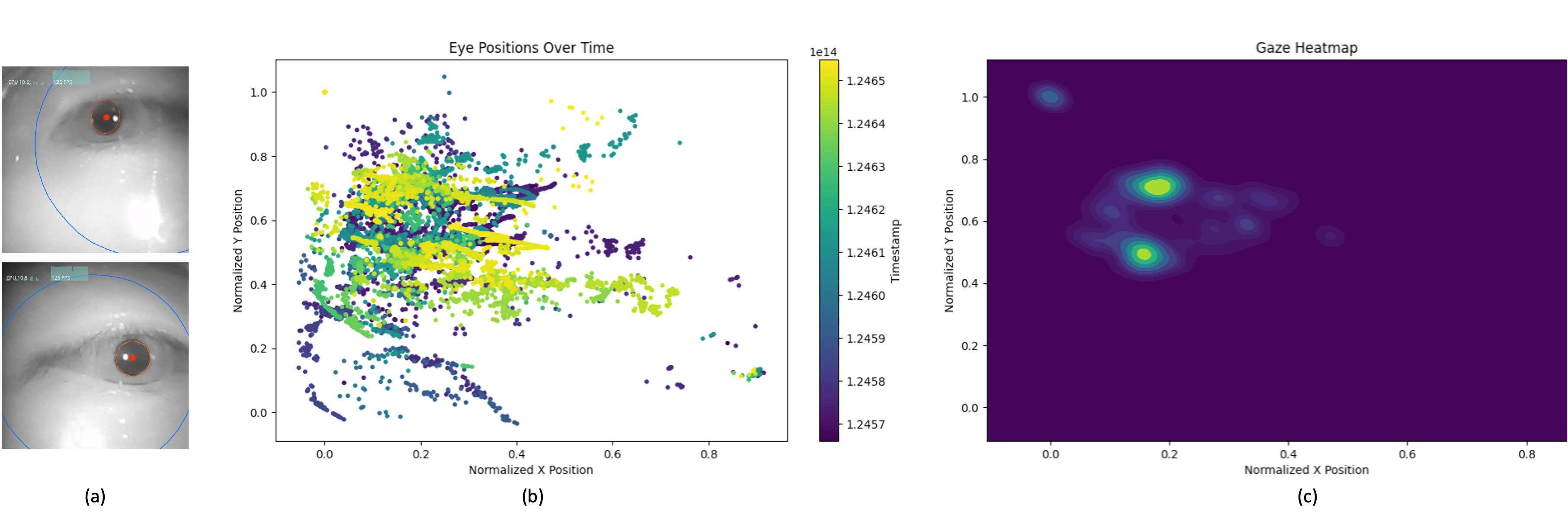}
\caption{(a) Dual-camera images captured by the head-mounted PupilLabs Core eye-tracker. (b) Temporal evolution of gaze positions. (c) Heatmap representing gaze distribution based on eye tracking data. }
\label{eyepos}
\end{figure*}

\subsubsection{Gesture Recognition Network}
Precise gesture recognition is paramount to facilitate seamless human-robot interaction within our HRC framework. It not only enables the identification of specific hand gestures but also provides insights into the user's status, such as hand occupation or openness for tool delivery. To accomplish this, we employ a specialized neural network tailored explicitly for gesture recognition. This network, structured as a fully connected feedforward neural network, takes as input the 21 landmark points representing the user's hand pose. These landmarks undergo meticulous processing to yield precise classifications, enabling us to discern the specific gesture being executed. This gesture recognition network serves as a cornerstone of our framework, enhancing the interpretation of user commands and overall interaction quality.

\subsubsection{Movement Recognition Network}
In addition to static hand gestures, recognizing human hand movements is crucial for achieving responsive human-robot collaboration, as it conveys dynamic information about the user's interaction preferences and other information. To address this need, we introduce a neural network architecture that combines Long Short-Term Memory (LSTM) and Fully Convolutional Network (FCN) layers for movement recognition. This network operates on sequences of 30 frames, adeptly capturing temporal dependencies and spatial features within the hand movement data. The network's output provides precise movement classifications, serving as a pivotal criterion for mode switching within our HRC framework. These specialized neural networks, governing both gesture and movement recognition, empower our system to comprehensively interpret and respond to user actions, ensuring a natural and intuitive collaborative experience.

\subsection{Dynamic Control Strategy Adaptation}
The approach involves switching between the Linear Quadratic Regulator (LQR) and the Proportional-Integral-Derivative (PID) controller based on real-time recognition of specific human hand movements using the Mediapipe framework and a trained neural network. The incorporation of this dynamic control strategy mode switch module is vital due to the distinct strengths and weaknesses of LQR and PID control algorithms. LQR excels in delivering precise, optimal control but struggles with dynamic changes and nonlinearities, while PID offers versatility and robustness but may lack the precision of LQR. By integrating both controllers and leveraging the control strategy mode switch module, our framework ensures seamless adaptation. The robot can switch between LQR for tasks demanding precision and PID for scenarios requiring adaptability, thereby maximizing performance across diverse HRC environments.

\subsubsection{Recognition of Human Hand Movements}
Dynamic control strategy adaptation relies on the real-time recognition of human hand movements, which serve as pivotal indicators for mode switching. The recognition process involves two key steps: utilizing the Mediapipe framework for extracting landmarks from real-time visual input, thereby capturing the nuances of hand movements, and employing a trained neural network for urgency classification. This neural network classifies observed hand movements, designating certain gestures, such as the ``give it to me" motion, as high urgency, necessitating a rapid transition to the more responsive PID control mode, while categorizing other gestures as low urgency, allowing the system to maintain precision in LQR mode for routine tracking tasks.

\subsubsection{Criteria for Mode Switching}
The criteria for switching between control strategies are intricately linked to the recognition of human hand movements:

\textit{High Urgency Gesture}: When the neural network classifies a hand movement as high urgency (e.g., ``give it to me" gesture), the system transitions from LQR to faster and more responsive PID control. This ensures rapid and accurate response to urgent requests.

\textit{Low Urgency or Medium Urgency Gesture}: When low or medium urgency gestures are detected, the system operates in LQR mode with different velocities. LQR provides precision and stability for routine tracking tasks.

\section{Experiment}
In this section, we provide an overview of the experimental setup designed to evaluate our proposed framework's performance. The primary aim of this experiment is to assess how effectively our framework facilitates seamless and intuitive interactions between humans and robots within a practical context.

The experiment was structured around a circuit repair task, carefully designed to comprehensively evaluate the capabilities of the HRC framework \footnote{The multimedia material is available at \url{https://sites.google.com/view/dhgfhma/home}}. The experiments were approved by the Ethics Committee of Imperial College London (21IC7042). Each participant was informed about the experiment's purpose and protocol and signed a consent form before the experiment. They began by using a soldering iron to heat a designated pad on the circuit, emulating a common electronics repair scenario. Subsequently, participants instructed the robot to deliver a desoldering pump, which they skillfully use to gently desolder a malfunctioning electronic component. Following the desoldering process, participants requested the robot to provide a soldering wire, which they employed to solder a new electronic component onto the circuit. Finally, participants asked for a wire cutter from the robot to trim the excess length of the components' legs, thereby completing the circuit repair task.

To gauge the performance of the HRC framework comprehensively, a set of performance metrics was collected. This includes the performance of each recognition network and robot positional error, which is used to assess the accuracy of the robot's movements, quantifying deviations along the x, y, and z-axes between the intended target positions and the actual positions of the robot's end-effector. In addition, human gaze analysis was conducted using the PupilLabs eye tracker to examine whether participant attention is diverted or distracted by the robot during the task.


\begin{figure}[t]
\centering
\includegraphics[width=1.0\linewidth]{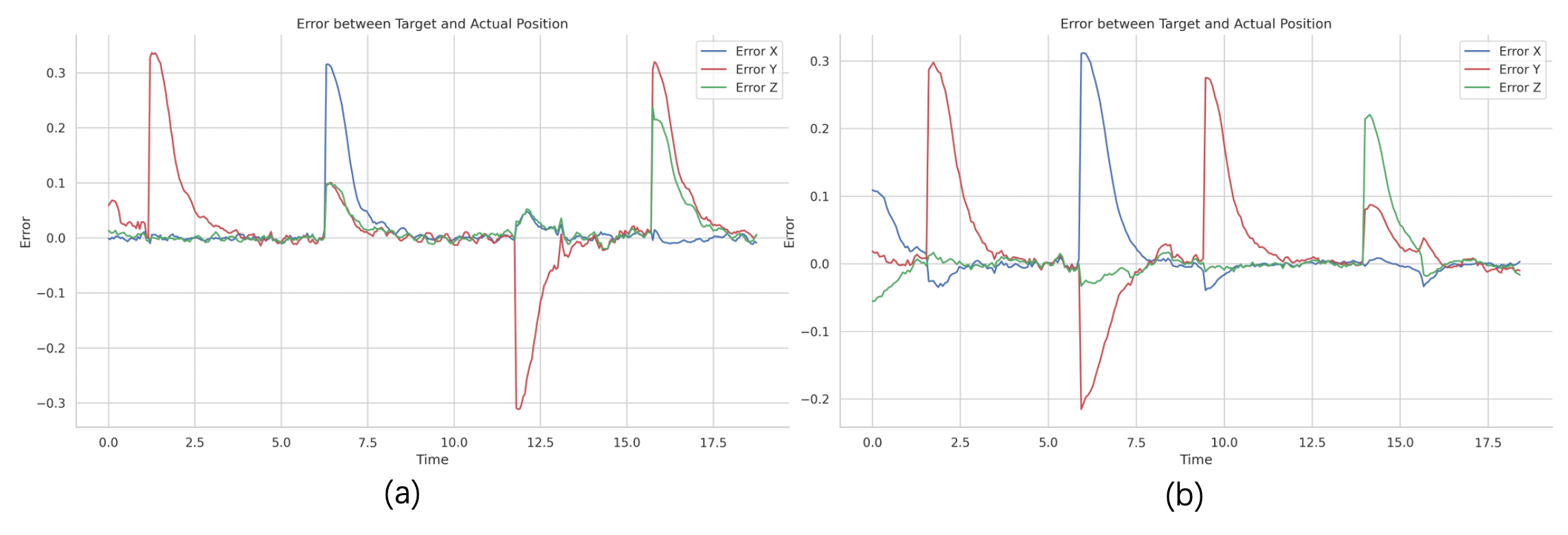}
\caption{Robot positional error relative to the target position over time: (a) LQR control with the state matrix to all zeros and the control input matrix as an identity matrix. (b) PID control with proportional gains set to 0.1, integral gains set to 0, and derivative gains set to 0.2 for all x, y, and z axes.}
\label{pos_error}
\end{figure}

In our experimental setup, we employed a 6 Degrees of Freedom (DoF) UR5e robot for human interaction, facilitated by the RealSense D435i camera, which captured 640 \(\times\) 480 RGB image frames for subsequent processing through the MediaPipe framework to collect landmark data. To ensure robust recognition, we curated a diverse dataset of pre-defined hand gestures and movements, encompassing variations in angles, camera positions, and lighting conditions. This dataset was diligently preprocessed, involving trimming and normalization, before training our recognition networks using PyTorch. Remarkably, our comprehensive framework operates seamlessly on an Intel i7-10510U CPU at a stable rate of 15Hz, ensuring an efficient and reliable interaction experience.

To ensure the safety of our experiment, we meticulously implemented several measures. A ``virtual wall" was established as a virtual boundary, confining the robot to a predefined workspace, preventing it from entering restricted or hazardous areas, and enhancing participant safety. A force/ torque sensor was also mounted on the robot's end-effector, serving as an immediate stop mechanism, swiftly halting the robot's motion upon any unexpected collisions, further bolstering participant safety. Furthermore, stringent limits on joint velocity and acceleration were enforced to mitigate the risk of abrupt or erratic robot movements.


To attain the intended system response of PID control, meticulous adjustment of the parameters \(K_p\), \(K_i\), and \(K_d\) is required. We set \(K_p\) to 0.1, \(K_i\) to 0, and \(K_d\) to 0.2 in the experiment, which was fine-tuned to strike a balance between swift response, precision in steady-state conditions, and overall system stability. It is worth noting that the PID control module serves as an alternative means for regulating the robot behavior, offering rapid response characteristics that are distinct from the LQR control module used in our system. The robot's response to the control strategies is illustrated in Fig. \ref{pos_error}. It can be observed that although LQR is smoother and more stable at the steady state, the fine-tuned PID is faster.

\section{Results and Discussion}
\begin{figure}[t]
\centering
\includegraphics[width=1.0\linewidth]{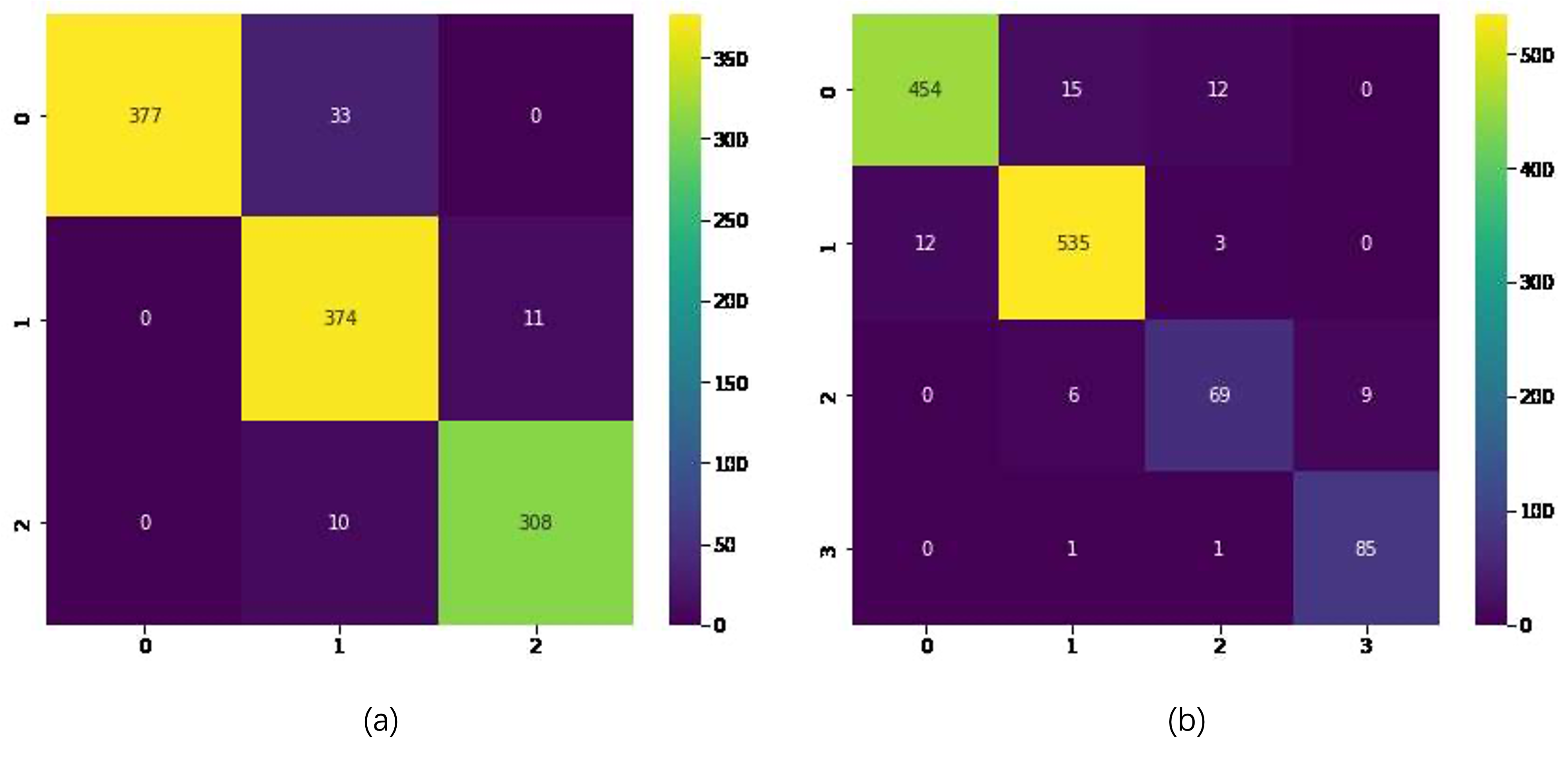}
\caption{Trained recognition neural network heatmap: (a) Gesture recognition network heatmap. (b) Movement recognition network heatmap.}
\label{nn_results}
\end{figure}

We present the outputs of the Gesture Recognition Network and Movement Recognition Network in the form of heatmaps in Fig.\, \ref{nn_results}. The Gesture Recognition Network heatmap illustrates the model proficiency in classifying hand gestures into three distinct categories: ``open," ``closed," and ``occupied." Each category is represented by a unique heatmap, showcasing the network's ability to discern and accurately categorize these gestures based on the provided hand landmark data. The Movement Recognition Network heatmap, on the other hand, highlights the network's effectiveness in classifying hand movements into four distinct urgency categories: representing low, medium, and high urgency, and one movement ``go away" that forces the robot back to the initial pose. These heatmaps provide insights into the network's capacity to interpret and classify dynamic hand movements, which is essential for real-time human intention recognition and mode switching within our HRC framework. The Gesture Recognition Network achieves an accuracy of 94.3\%, while the Movement Recognition Network reaches an impressive accuracy of 97.6\%. 

In our experiment, we evaluated mean pupil diameter, blink frequency, and fixation frequency to assess human workload, as established in  \cite{guo2021eye}. Minor changes were observed in blink frequency, indicating that the robot intervention did not significantly distract from human intentions. However, we observed changes in pupil diameter, which increased from 2.54mm to 3.09mm, and blink frequency, which increased from 0.198 to 0.343 per second. These alterations suggest a moderate increase in human mental load due to the robot intervention, but overall, the impact on human intention and workload remained relatively low during the task.

Fig. \ref{eyepos}(a) depicts the temporal evolution of gaze positions over the course of the study. Each data point corresponds to the normalized gaze position of a participant, with the X-axis representing horizontal gaze coordinates and the Y-axis representing vertical gaze coordinates. Fig. \ref{eyepos}(b) is the heatmap generated from eye tracking data. The figure utilized colormap to illustrate the participant's gaze behavior throughout the experimental session and the density of gaze points across the screen. The heatmap provides valuable insights into the areas of interest and gaze distribution throughout the experimental task, shedding light on participant visual attention patterns. Analysis of gaze position trajectories enables a deeper understanding of how visual attention evolves in response to the robot movement and tasks.

To assess the performance of our framework, we conducted experiments under three primary conditions: with robot delivery (where the robot follows human hands and delivers objects), without robot delivery (the robot delivers objects to a fixed location upon voice command), and our framework without hand following and voice recognition (requiring users to fetch objects themselves), as shown in Table \ref{results_table}. Notably, when hand following was incorporated, allowing the robot to adapt its movements based on the user's hand position and gestures, led to a significant reduction in the average task completion time. This result underscores the significance of hand following in streamlining interactions. Conversely, when both hand following and voice recognition were removed, placing the onus on users to fetch tools themselves, the completion time slightly increased. These findings indicate that the framework's adaptability and intuitiveness contribute to more efficient HRC. To gain a more comprehensive understanding of its performance, future experiments will explore additional metrics user satisfaction, and a comprehensive mental load analysis.


\begin{table}[ht]
\centering
\captionsetup{width=0.8\linewidth}
\caption{Experimental results: average task completion time(s) under different conditions}
\begin{tabularx}{\linewidth}{@{}>{\centering\arraybackslash}X >{\centering\arraybackslash}X@{}}
\toprule
{Conditions} & {Avg Time (s)} \\
\midrule
w/o robot delivery & 367 \\
w robot delivery & \textbf{289} \\
w/o robot delivery and voice recognition & 392 \\
\bottomrule
\end{tabularx}
\label{results_table}
\end{table}

It is important to acknowledge that this study was conducted with a relatively small number of participants, and while the results are promising, further validation on a larger and more diverse sample is required to ensure the generalizability and feasibility of our system. However, it's essential to highlight that our experiment required a certain level of expertise in normal task execution. Consequently, considerable time and effort were invested in training the operators to perform the tasks effectively. This level of specialization could pose challenges when recruiting subjects for similar experiments. 

Additionally, the choice of objects used in the experiment was somewhat constrained to match the experimental setup, which may not fully represent the diversity of real-world scenarios. The primary focus of this work was to demonstrate the feasibility of our system through a demonstration-based approach. When considering tasks that require more generic applicability, such as object grasping, it becomes necessary to tailor the system to different gripper shapes and object recognition strategies, which extends beyond the scope of this study.


\section*{Acknowledgment}
This work was supported by Lloyd’s Register Foundation, a charitable foundation, helping to protect life and property by supporting engineering-related education, public engagement and the application of research. 
\href{www.lrfoundation.org.uk}{www.lrfoundation.org.uk}.
We thank Darren Williams for helpful discussion and valuable comments on the manuscript.
\bibliographystyle{unsrt}
\bibliography{thesis}

\end{document}